%% file: wacv_paper.tex
\documentclass[10pt,twocolumn,letterpaper]{article}

\usepackage[pagenumbers]{wacv} 

\usepackage{graphicx}
\usepackage{amsmath}
\usepackage{amssymb}
\usepackage{booktabs}

\usepackage{lipsum}
\usepackage{multirow, multicol}
\usepackage{array}
\newcolumntype{L}[1]{>{\raggedright\let\newline\\\arraybackslash\hspace{0pt}}m{#1}}
\newcolumntype{C}[1]{>{\centering\let\newline\\\arraybackslash\hspace{0pt}}m{#1}}
\newcolumntype{R}[1]{>{\raggedleft\let\newline\\\arraybackslash\hspace{0pt}}m{#1}}
\usepackage{caption}
\usepackage{subcaption}
\usepackage{pifont}
\newcommand{\cmark}{\ding{51}}%
\newcommand{\xmark}{\ding{55}}%

\newcommand{\mycomment}[1]{}
\usepackage{enumitem}
\usepackage{lipsum}
\usepackage{xcolor}

\usepackage[pagebackref,breaklinks,colorlinks]{hyperref}

\usepackage[capitalize]{cleveref}
\crefname{section}{Sec.}{Secs.}
\Crefname{section}{Section}{Sections}
\Crefname{table}{Table}{Tables}
\crefname{table}{Tab.}{Tabs.}

\def\wacvPaperID{***} 
\def\confName{WACV}
\def\confYear{2025}

\begin{document}

\title{Unsupervised Video Highlight Detection by Learning from\\ Audio and Visual Recurrence}


\author{Zahidul Islam$^{1}$ \qquad\qquad Sujoy Paul$^{2}$ \qquad\qquad Mrigank Rochan$^{1}$ \\
	$^1$University of Saskatchewan, Canada \qquad $^2$Google DeepMind
}

\maketitle

\begin{abstract}
  \input texs/abstract.tex
\end{abstract}

\input texs/intro.tex
\input texs/related.tex
\input texs/approach.tex
\input texs/experiments_wacv.tex
\input texs/conclusion.tex

{\small
\bibliographystyle{ieee_fullname}
\bibliography{highlight_refs}
}

\end{document}

%% file: texs/abstract.tex
With the exponential growth of video content, the need for automated video highlight detection to extract key moments or highlights from lengthy videos has become increasingly pressing. This technology has the potential to enhance user experiences by allowing quick access to relevant content across diverse domains. Existing methods typically rely either on expensive manually labeled frame-level annotations, or on a large external dataset of videos for weak supervision through category information. To overcome this, we focus on unsupervised video highlight detection, eliminating the need for manual annotations. We propose a novel unsupervised approach which capitalizes on the premise that significant moments tend to recur across multiple videos of the similar category in both audio and visual modalities. Surprisingly, audio remains under-explored, especially in unsupervised algorithms, despite its potential to detect key moments. Through a clustering technique, we identify pseudo-categories of videos and compute audio pseudo-highlight scores for each video by measuring the similarities of audio features among audio clips of all the videos within each pseudo-category. Similarly, we also compute visual pseudo-highlight scores for each video using visual features. Then, we combine audio and visual pseudo-highlights to create the audio-visual pseudo ground-truth highlight of each video for training an audio-visual highlight detection network. Extensive experiments and ablation studies on three benchmarks showcase the superior performance of our method over prior work.

%% file: texs/intro.tex
\section{Introduction}\label{sec:intro}
Video highlight detection is a critical task in the realm of video content analysis, where the aim is to automatically identify and extract the most important or engaging segments from lengthy video content \cite{badamdorj2021joint, badamdorj2022contrastive, hong2020mini, xiong2019lessLM}. As the volume of video data on the internet continues to surge, there is a growing demand for efficient methods to navigate and consume such content. With applications ranging from sports broadcasting to content creation, education, marketing, surveillance, entertainment, and beyond, video highlight detection holds significant promise. It can notably enhance user experience in sports broadcasts, aid content creators in generating engaging trailers or highlight reels, and offer substantial benefits across various domains.

\begin{figure*}[htb]
     \centering
     \begin{subfigure}[b]{0.43\textwidth}
         \centering
         \includegraphics[width=\textwidth]{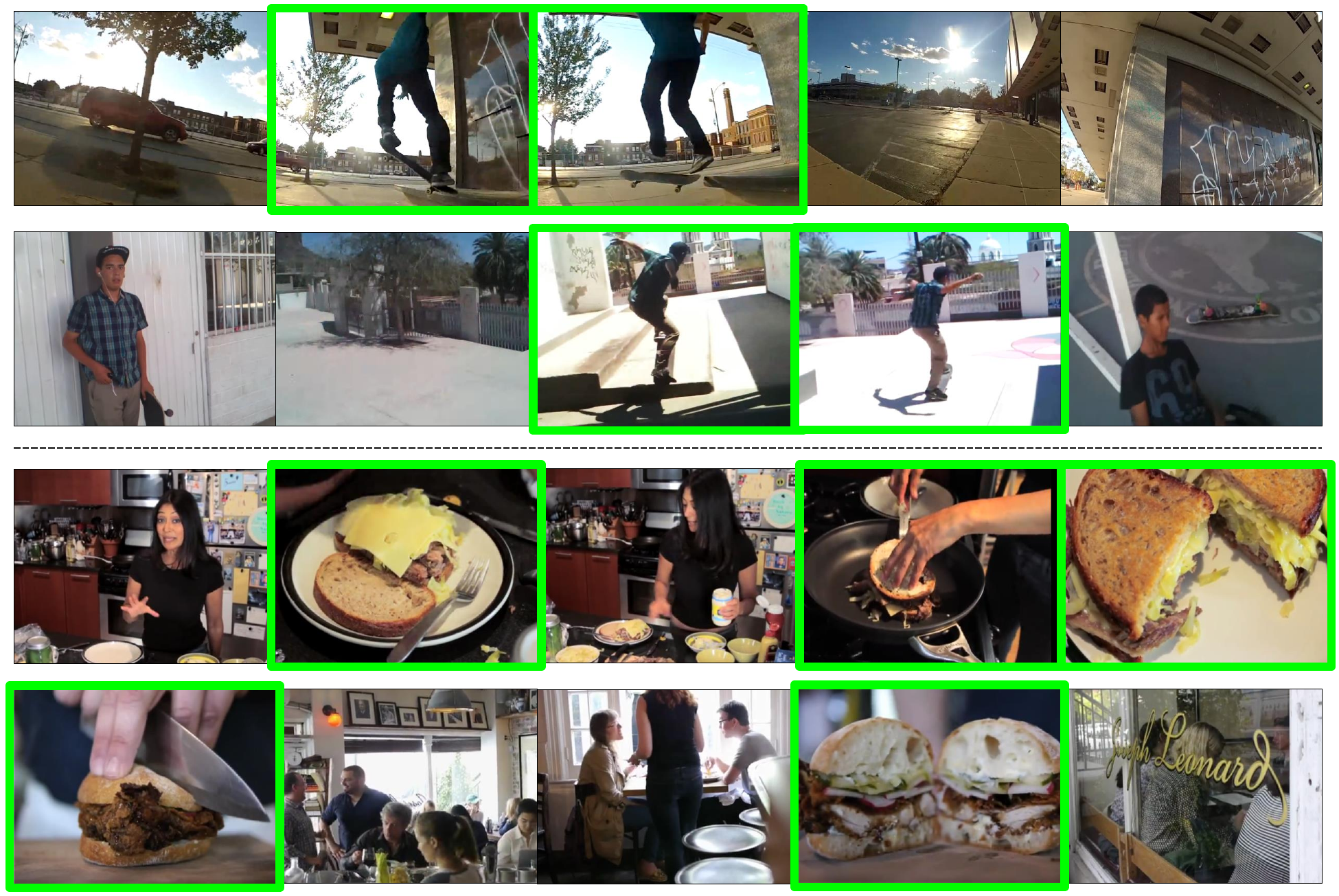}
         \caption{}
         \label{fig:int_a}
     \end{subfigure}
     \hfill
     \begin{subfigure}[b]{0.43\textwidth}
         \centering
         \includegraphics[width=\textwidth]{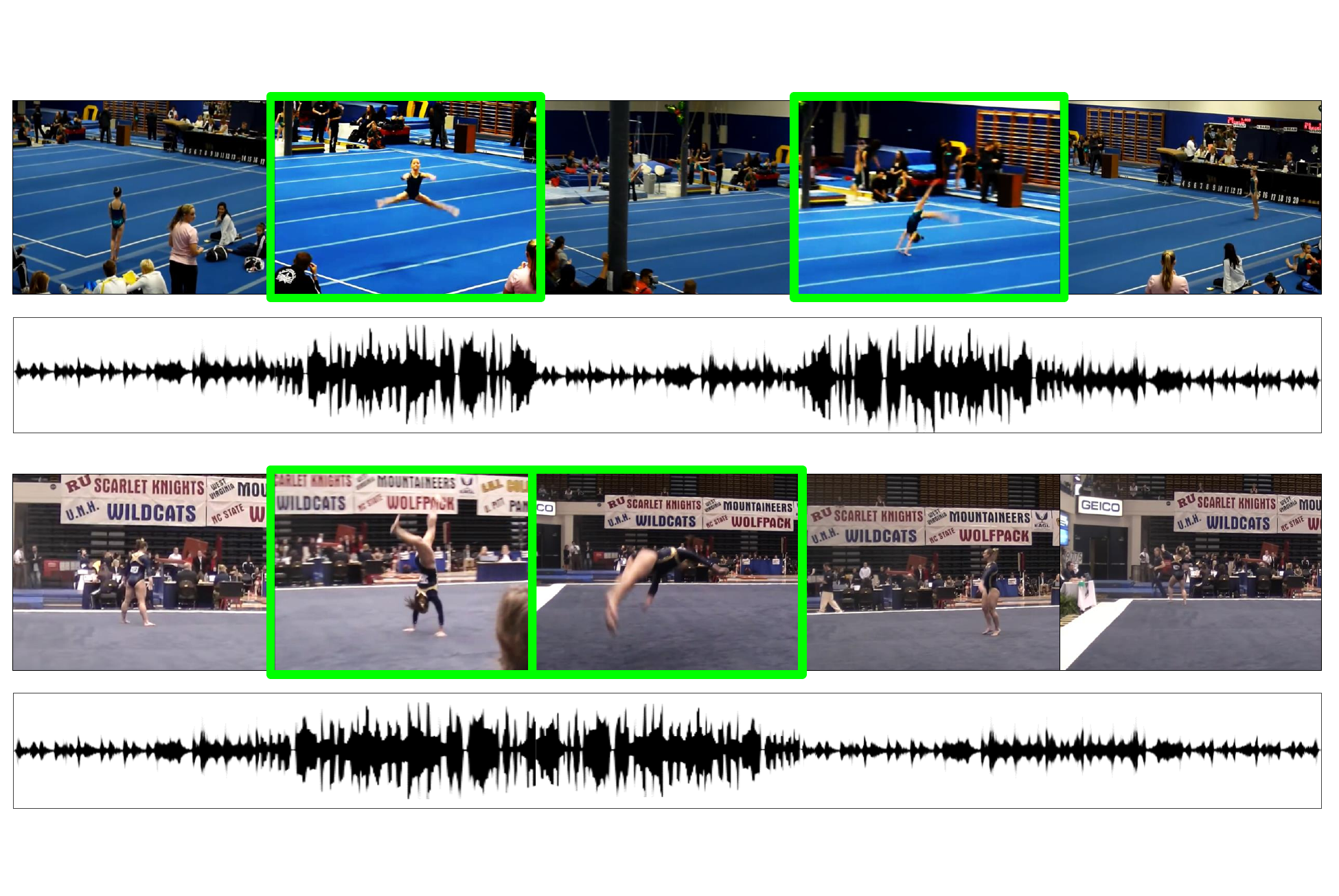}
         \caption{}
         \label{fig:int_b}
     \end{subfigure}
     \hfill
        \caption{a) \textit{Visual Recurrence:} Highlight of \textit{skating} videos mostly consist of jump tricks, which appear frequently in multiple videos (first and second rows). Similarly, in \textit{cooking} videos, close-up shots of food depicting various actions, such as chopping and pan-frying, are commonly appearing highlight moments (third and fourth rows). b) \textit{Audio Recurrence:} In \textit{gymnastics} videos, loud cheers and claps are recurring audio cues, which occur when the spectators react to interesting and highlight-worthy moves such as flips or cartwheels. Note that the highlight clips marked in green are also the annotated ground-truth highlights of the example videos in the benchmark datasets.}
        \label{fig:intuition}
\end{figure*}

Supervised approaches to video highlight detection are popular but face challenges due to the requirement of expensive manually annotated frame-level supervision \cite{badamdorj2021joint, jiao2018deepranking, wang2020learningtrailer, yu2018deep, gygli2016video2gif, jiao2018deepranking, rochan2020adaptive, liu2022umt}. To address this issue, there is a stream of research in weakly supervised learning \cite{xiong2019lessLM, ye2021temporal,cai2018weaklyVESD, hong2020mini, panda2017weaklyDSN}, which utilizes video-level labels such as video category as a weak supervision signal. However, these approaches typically require a large external dataset, such as web-crawled videos, for model training. In light of this, the largely unexplored development of unsupervised algorithms presents a promising avenue for addressing the need for automated highlight detection without requiring any labels. In this paper, we introduce an innovative unsupervised method for video highlight detection. Our approach leverages cues from both audio and visual components of the video to improve video highlight detection. Interestingly, audio cues are often overlooked, but they can be highly informative for highlight detection. To the best of our knowledge, our work represents the first attempt to exploit both audio and visual components in unsupervised learning for video highlight detection without requiring any large external dataset.

Videos with similar content or actions, tend to exhibit recurrence of key moments in both the audio and visual modalities. By \textit{recurrence}, we mean the repetition of specific patterns or features in multiple videos of similar categories. These may manifest as audio cues, like the repeated occurrence of specific sounds or phrases, or as visual cues, such as the reappearance of particular objects or scenes. For example, in cooking videos, close-up shots of food depicting certain actions appear frequently, such as chopping vegetables, stirring ingredients, or the sizzling sound of food being cooked. These recurring visual and audio elements can serve as strong indicators of important moments in the video. Similarly, in sports videos, the cheering of the crowd or the excitement in the commentator's voice may recur as audio cues, while the slow-motion replay of a goal or a player's celebration may recur as visual cues, both signaling key moments for highlight. We also visualize these observations through some examples in Figure \ref{fig:intuition}. Hence, we combine the strengths of both auditory and visual cues to detect highlights through their inherent recurrence.

We devise an unsupervised algorithm to identify these recurrences and obtain a supervisory signal in the form of audio-visual pseudo-highlight to train a highlight detection network. In practice, to circumvent the need for any manual annotation, we first use a clustering technique to identify pseudo-categories of videos in the dataset. Then, we segment a video into clips and compare each clip with all the clips across videos of the same pseudo-category using their audio and visual features to obtain \textit{audio pseudo-highlight} scores and \textit{visual pseudo-highlight} scores, respectively. We aggregate these audio and visual pseudo-highlight scores to obtain the \textit{audio-visual pseudo-highlight} scores for the clips in the video and select top-scored clips to compile the audio-visual pseudo-highlight of the video which we use to train our audio-visual highlight detection network. This network, partly inspired by \cite{badamdorj2021joint}, comprises unimodal and bimodal attention mechanisms. The unimodal attention mechanism captures relationships between the video clips based on the features of the same modality, whereas the bimodal attention mechanism focuses on the interrelationship between the two modalities. Our network is trained end-to-end using the audio-visual pseudo highlight. At test time, we concatenate the top-scoring clips of an input video to generate its highlight. The proposed method not only outperforms state-of-the-art unsupervised methods but also demonstrates comparable or superior performance to state-of-the-art weakly supervised methods.

In summary, our unsupervised video highlight detection method, which integrates audio and visual cues based on the recurrence of patterns or features, offers a novel approach for highlight detection. By capitalizing on recurring features in audio and visual modalities, we generate audio-visual pseudo-highlights for training the detection network, replacing costly manual annotations. Our highlight detection model surpasses prior work and strong baselines on three standard benchmark datasets. It also contributes to the growing demand for automated and efficient methods for video highlight detection. Additionally, it paves the way for future research in unsupervised learning using audio and visual modalities for complex video understanding tasks.

%% file: texs/related.tex
\section{Related Work}\label{sec:related}
{\flushleft\textbf{Video Highlight Detection:}} Most of the existing approaches for highlight detection rely on manually annotated frame-level supervision \cite{gygli2016video2gif, jiao2018deepranking, rochan2020adaptive, wang2020learningtrailer, yu2018deep, badamdorj2021joint, liu2022umt}. However, these annotations are laborious and expensive to obtain. To address this limitation, some works focus on using only video-level tags or category information as weak supervision \cite{yang2015unsupervisedRRAE, cai2018weaklyVESD, hong2020mini, panda2017weaklyDSN,xiong2019lessLM, ye2021temporal}. Several works adopt ranking frameworks. For example, LM \cite{xiong2019lessLM} exploits video duration and ranks clips from shorter videos higher, whereas, Ye \textit{et al.} \cite{ye2021temporal} leverages temporal reasoning and encodes cross-modal relationships using an efficient audio-visual tensor fusion mechanism. However, these methods require training on large-scale external data. A recent unsupervised method based on contrastive learning \cite{badamdorj2022contrastive} does not utilize any external data. However, they do not exploit audio for unsupervised learning. Some recent works explore supervised query-based highlight detection, where the goal is to extract highlight relevant to a given textual query \cite{lei_moment_detr, liu2022umt, moon2023query}. In contrast, we focus on unsupervised audio-visual highlight detection which does not rely on any text input.

{\textbf{Video Summarization:}} When summarizing videos, the aim is to generate a coherent and concise synopsis of the video, whereas video highlight detection aims to extract significant moments. Earlier works on video summarization utilize unsupervised heuristics such as representativeness and diversity of the selected clips \cite{khosla2013large, kim2014reconstructing, song2016beautythumb, lee2012discovering, mahasseni2017unsupervised, panda2017collaborative, song2015tvsum, zhou2018deepdiversity}. Some methods use only weak video-level supervision \cite{cai2018weaklyVESD, panda2017weaklyDSN, potapov2014category, rochan2019videounpaired}, while others employ supervised learning \cite{gong2014diverse, gygli2014creating, gygli2015videosubmodular, rochan2018videoFCSN, zhang2016summary, zhang2016videolongshort, zhao2017hierarchical}. Recent works focus on capturing contextual dependencies using recurrent networks \cite{zhang2016videolongshort, zhao2017hierarchical} or attention layers \cite{fajtl2019summarizingattention, summarizationSTVT}. Our work is partly related to a recent study on instructional video summarization \cite{narasimhan2022tldw}, which uses visual data and textual transcripts to construct pseudo-summaries comprised of the most salient steps to train their model. However, they do not exploit audio modality and require category information of videos for weak supervision.

%% file: texs/approach.tex

\begin{figure*}[t] 
\centering
\includegraphics[width=0.88\textwidth]{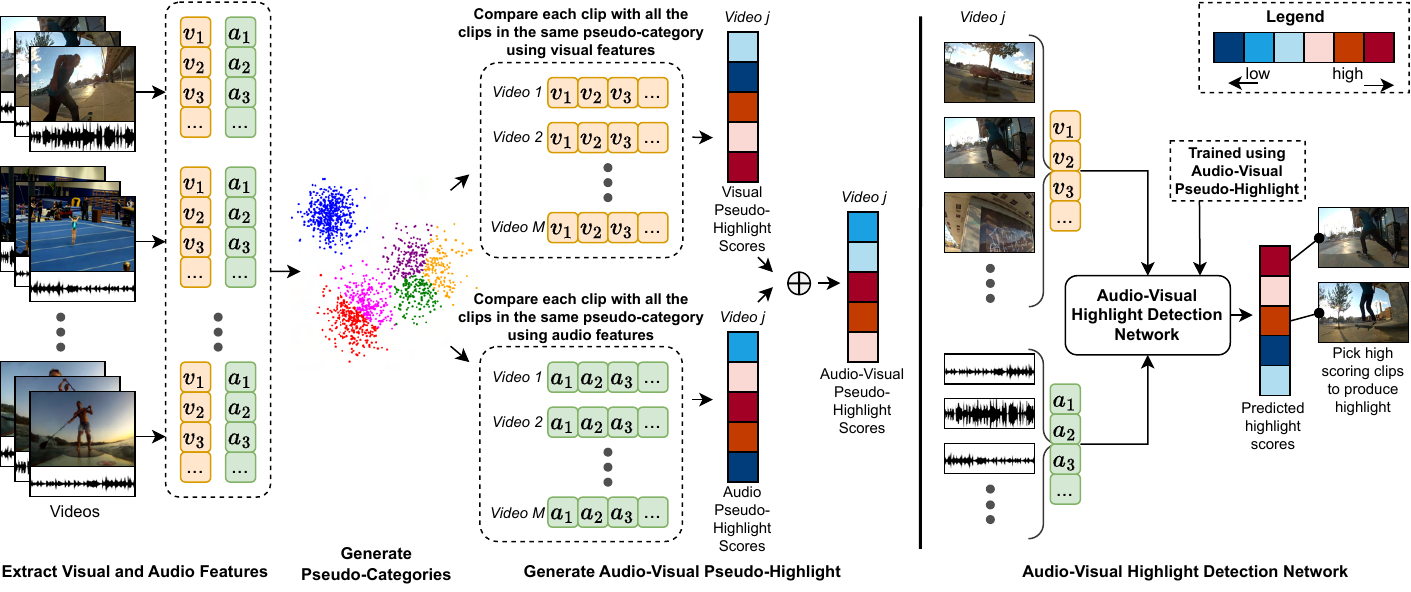} 
\caption{ An overview of our unsupervised highlight detection framework. We first extract visual and audio features from each clip of a video. We then use a clustering technique to identify pseudo-categories of the videos. Next, we compare each clip of a video with all the clips across videos of the same pseudo-category using both audio and visual features to obtain audio-visual pseudo-highlight (AV-PH) of the video. Using the audio-visual pseudo-highlight as supervision, we train our audio-visual highlight detection network to assign a highlight score to each video clip. We pick the high scoring clips to obtain the highlight of the video.}
\label{fig:method}
\end{figure*}


\section{Our Approach}\label{sec:approach}
We introduce a novel unsupervised highlight detection approach that utilizes both audio and visual modalities to extract highlights from a video without requiring any manual annotations. Our highlight detection method builds upon the notion that in a group of similar category videos, audio and visual cues associated with highlight moments, such as cheering or clapping in sports videos, tend to recur or repeat across multiple videos. Broadly, our approach consists of three steps. Firstly, we group similar videos into pseudo-categories using a clustering-based algorithm. Next, we calculate an audio-visual pseudo-highlight score for each clip of a video based on its recurrence in the audio and visual modalities using a similarity measure, indicating its likelihood of being a highlight. Finally, we select the clips with high scores to generate the audio-visual pseudo-highlight for each video and use it as ground-truth to train our audio-visual highlight detection network. Figure \ref{fig:method} shows an overview of our framework.

Let's say we have a set of $M$ unlabeled videos $\{V_j\}_{j=1}^{M}$. We split each video $V_j$ into clips with an equal number of frames, resulting in $n$ clips. From each clip $c_i$ (where $i=1,2,...n$), we extract its corresponding visual features $v_i \in \mathbb{R}^{d_v}$ using a pre-trained visual feature extractor and audio features $a_i \in \mathbb{R}^{d_a}$ using a pre-trained audio feature extractor. Each video $V_j$ can then be represented using its corresponding set of visual features, $\{v_i\}_{i=1}^n$, and audio features, $\{a_i\}_{i=1}^n$. The aim of our highlight detection method is to predict a set of highlight scores $\{h_i\}_{i=1}^n$, where $h_i$ indicates the highlight score of each clip $c_i$ in video $V_j$. In the following, we discuss the three broad steps in our approach.

\subsection{Generating Pseudo-Categories} \label{subsec:gpc}
In this step, we aim to assign the unlabeled videos to multiple groups. We refer to these groups as \textit{pseudo-categories}. We adopt a clustering-based approach to find the pseudo-category of each video. From each video $V_j$ in the training set of a dataset, we calculate the mean of its clip-level visual features $\{v_i\}_{i=1}^n$ as $\bar{v} = \frac{1}{n} \sum_{i=1}^{n} v_i$. Similarly, we calculate the mean of its clip-level audio features $\{a_i\}_{i=1}^n$ as $\bar{a}=\frac{1}{n} \sum_{i=1}^{n} a_i$. Subsequently, we concatenate $\bar{v}$ and $\bar{a}$ to obtain a video-level audio-visual feature representation $\bar{f} = [\bar{v}; \bar{a}]$ of the video, which we utilize for clustering. We reduce the dimensionality of $\bar{f}$ using UMAP \cite{mcinnes2018umap}, resulting in transformed $\bar{f} \in \mathbb{R}^{10}$, which is a common pre-processing step in clustering literature \cite{ronen2022deepdpm, mcconville2021n2d}. Following standard practice for finding an optimal number of clusters, we iteratively apply the $K$-means clustering algorithm using the transformed features $\bar{f}$ of the videos for a range of values for $K$. For each $K$, we calculate a standard unsupervised clustering fitness metric, the Silhouette Coefficient (SC) \cite{rousseeuw1987silhouettes}, and choose the $K$ with the highest SC as the optimal number of clusters. We use the cluster labels assigned by clustering with the optimal $K$ as our pseudo-categories for the videos. These pseudo-categories are generated by clustering video-level audio-visual features extracted from pre-trained classification models, potentially grouping together videos with similar semantics. Next, we utilize these pseudo-categories when generating the audio-visual pseudo-highlight for each video.

\subsection{Generating Audio-Visual Pseudo-Highlight}
We generate an audio-visual pseudo-highlight for each video to train our model. For each video, we first compare each of its clips with all the clips across videos of the same pseudo-category using their audio and visual features to obtain audio pseudo-highlight and visual pseudo-highlight scores, respectively. We then aggregate these scores to generate audio-visual pseudo-highlight of the video.

{\textbf{Audio Pseudo-Highlight (A-PH):}} Let, there are $K$ videos assigned to a particular pseudo-category, and $\{a_k\}_{k=1}^S$ represents a set of audio features corresponding to all the $S$ clips across all $K$ videos of that pseudo-category. We assign an audio pseudo-highlight score $aph_i$ to the $i$-th clip of a video based on how repetitive the corresponding audio features $a_i$ are in the videos of that pseudo-category (referred to as audio recurrence). To achieve this, we utilize cosine similarity to compare the audio features of a clip $a_i$ in a video with all the clips in the pseudo-category. We can write $aph_i$ computation as follows:
\begin{equation}
    \label{eq:aph}
    aph_i = \frac{1}{S} \sum_{k=1}^S \frac{ a_i \cdot a_k}{\|{a_i}\| \|{a_k}\|}
\end{equation}

{\textbf{Visual Pseudo-Highlight (V-PH):}} Let's consider $\{v_k\}_{k=1}^S$ as a set of visual features of all the $S$ clips belonging to all $K$ videos of a pseudo-category. Similar to the process of computing audio pseudo-highlight scores, we compute a visual pseudo-highlight score $vph_i$ for the $i^{th}$ clip of a video based on the similarity of the corresponding visual features $v_i$ in that pseudo-category (referred to as visual recurrence). We use cosine similarity to compare the visual features of a clip $v_i$ of a video with all the clips belonging to the pseudo-category. We can write $vph_i$ computation as:
\begin{equation}
    \label{eq:vph}
    vph_i = \frac{1}{S} \sum_{k=1}^S \frac{ v_i \cdot v_k}{\|{v_i}\| \|{v_k}\|}
\end{equation}

{\textbf{Audio-Visual Pseudo-Highlight (AV-PH):}} For each clip in a video, we first compute the average of audio pseudo-highlight and visual pseudo-highlight scores. Then, we select the top $t\%$ clips based on the average scores to obtain the audio-visual pseudo-highlight (AV-PH) of the video, which we use as supervisory signal for training our network.

\subsection{Audio-Visual Highlight Detection Network}
We adopt the network architecture from prior work \cite{badamdorj2021joint} for our audio-visual highlight detection network (AV), a network design that has been proven to be highly effective in supervised highlight detection. At its core, our network initially employs unimodal self-attention layers \cite{attentionvaswani} to capture clip-level temporal relationships within each modality using their features. Subsequently, these self-attended visual and audio features are fed into bimodal cross-attention layers to encode cross-modal dependencies and produce bimodal attended features. Finally, the self-attended features and bimodal attended features are combined and forwarded to fully-connected layers to predict the highlight score of each clip in the video.

More concretely, given a video with $n$ clips, our audio-visual model processes the clip-level visual features $\{v_i\}_{i=1}^n$ using a self-attention layer $\text{Attn}_{v \rightarrow v}$ and the clip-level audio features $\{a_i\}_{i=1}^n$ using another self-attention layer $\text{Attn}_{a \rightarrow a}$. Then, two bimodal attention layers $\text{Attn}_{v \rightarrow a}$ and $\text{Attn}_{a \rightarrow v}$ process the self-attended visual features $\{v_i^v\}_{i=1}^n$ and self-attended audio features $\{a_i^a\}_{i=1}^n$ to produce bimodal attended features, $\{v_i^a\}_{i=1}^n$ and $\{a_i^v\}_{i=1}^n$, respectively. Finally, a score regressor module (SR) combines self-attended and bimodal attended features using learnable weights and passes them through two fully-connected layers to predict the highlight score ${h_i}$ for each clip in the video. These operations can be expressed as follows:
\begin{align} 
\small
  \label{eq:model1}
  \{v_i^v\}_{i=1}^n = \text{Attn}_{v \rightarrow v}(\{v_i\}_{i=1}^n)\\
  \label{eq:model2}
  \{a_i^a\}_{i=1}^n = \text{Attn}_{a \rightarrow a}(\{a_i\}_{i=1}^n)\\
  \label{eq:model3}
  \{a_i^v\}_{i=1}^n = \text{Attn}_{a \rightarrow v}(\{a_i^a\}_{i=1}^n, \{v_i^v\}_{i=1}^n)\\
  \label{eq:model4}
  \{v_i^a\}_{i=1}^n = \text{Attn}_{v \rightarrow a}(\{v_i^v\}_{i=1}^n, \{a_i^a\}_{i=1}^n)\\
  \label{eq:model5}
  \{h_i\}_{i=1}^n = \text{SR}(\{v_i^v\}_{i=1}^n,\{a_i^a\}_{i=1}^n,\{v_i^a\}_{i=1}^n,\{a_i^v\}_{i=1}^n)
\end{align} 

The training procedure for our network follows the standard approach used in the supervised network \cite{badamdorj2021joint}. However, instead of using ground-truth annotations, we use audio-visual pseudo-highlights as the supervisory signal to train our network. We optimize our network by minimizing the binary cross-entropy loss between the predicted highlight scores and the audio-visual pseudo-highlights (AV-PH).

%% file: texs/experiments_wacv.tex
\section{Experiments}\label{sec:exp}
\subsection{Datasets and Settings} \label{subsec:exp_setup}
We evaluate our method using three benchmark highlight detection datasets: YouTube Highlights \cite{sun2014rankingdomainspecific}, TVSum \cite{song2015tvsum}, and QVHighlights \cite{lei_moment_detr}. YouTube Highlights is constructed by mining YouTube videos related to six specific categories such as parkour, gymnastics, skiing, and so on, with about 100 videos in each category. We utilize the train and test splits provided in this dataset. TVSum has 50 videos across 10 diverse categories. Following prior works \cite{badamdorj2021joint, badamdorj2022contrastive, rochan2019videounpaired}, we randomly split this dataset with $80\%$ of the videos for training and $20\%$ for testing, and we run our experiments on this dataset five times and report the average performance. QVHighlights is larger with over 10,000 videos. It is primarily designed for query-focused video highlight detection and moment retrieval. Each video is associated with a textual query and corresponding saliency/highlight scores. The dataset comes with standard train, validation, and test splits with a ratio of 70:15:15. Since our method only requires videos, we ignore the user query annotations. For a fair comparison, on QVHighlights, we evaluate our method against prior non-query-based methods.

{\textbf{Features:}} For YouTube and TVSum, we follow prior work \cite{badamdorj2022contrastive, badamdorj2021joint} and use a 3D-CNN comprised of a ResNet-34 backbone to extract visual features from each clip. For QVHighlights dataset, following \cite{lei_moment_detr, liu2022umt}, we extract visual features using SlowFast \cite{feichtenhofer2019slowfast} and video encoder of CLIP (ViT-B/32) \cite{clip_radford2021learning}. For all datasets, we employ PANN \cite{kong2020panns} audio network pre-trained on AudioSet \cite{audioset} to extract audio features. 

{\textbf{Evaluation Metrics:}} Following prior works on QVHighlights \cite{liu2022umt, lei_moment_detr}, we report our performance using Mean Average Precision (mAP), which considers the highlight scores for all the clips, and HIT@1, which considers the hit ratio of the clip with the highest score for each video. We consider only the clips rated as \textit{Very Good} by users for evaluation. On YouTube, we evaluate using mAP, and on TVSum, we report mAP on the top five predicted clips (top-5 mAP) as in prior works \cite{badamdorj2022contrastive, badamdorj2021joint}. All metrics are reported as percentages. 

{\textbf{Implementation Details:}} We implement our models using PyTorch \cite{pytorch}. For YouTube and TVSum, we utilize one self-attention and one bi-modal attention module in our audio-visual highlight detector network, as in the previous method \cite{badamdorj2021joint}. However, since QVHighlights is a much larger dataset, following previous studies \cite{liu2022umt}, we introduce an additional self-attention module and fully connected layer in the score regressor to handle the increased complexity and scale of the data. We use Adam to optimize \cite{kingma2014adam} our models. We train our models for 20 epochs with a learning rate of \(2.5\times10^{-3}\) on TVSum and for 100 epochs with a learning rate of \(5\times10^{-3}\) on YouTube. For QVHighlights, we train our models with a learning rate of \(5\times10^{-4}\) for $10$ epochs. As mentioned in Sec. \ref{subsec:gpc}, we empirically select the number of clusters, $K$, by maximizing the Silhouette Coefficient. For all three datasets, we search in the range of 4 to 15 to find the value of $K$. For YouTube, TVSum, and QVHighlights, we find the optimal values of $K$ to be 6, 8, and 7, respectively. Following prior works \cite{badamdorj2022contrastive, hong2020mini}, we select the top $t=50\%$ clips based on the audio-visual pseudo-highlight scores (AV-PH) to create pseudo-highlights for training.

{\textbf{Comparison Methods:}} We compare the performance of our approach on the three datasets with state-of-the-art weakly supervised methods which require category or topic information, including TC \cite{ye2021temporal}, MN \cite{hong2020mini}, LM-A \cite{xiong2019lessLM}, LM-S \cite{xiong2019lessLM}, RRAE \cite{yang2015unsupervisedRRAE}, MBF \cite{chu2015videoMBF}, CVS \cite{panda2017collaborative}, DSN \cite{panda2017weaklyDSN}, and VESD \cite{cai2018weaklyVESD}. We also compare with prior unsupervised approaches, including SG \cite{mahasseni2017unsupervised}, BT \cite{song2016beautythumb}, CHD \cite{badamdorj2022contrastive}, and MT \cite{li2024highlight}. Among the compared methods, MN, TC, and MT utilize both visual and audio modalities. On QVHighlights, we compare with prior unsupervised approaches that do not utilize textual queries: BT \cite{song2016beautythumb} and CHD \cite{badamdorj2022contrastive}.

\subsection{Highlight Detection Results}

{\textbf{YouTube:}} In Table \ref{table:comparison_YouTube}, we compare the performance of our approach on YouTube. Among the compared methods, CHD is the only method that is fully unsupervised and does not require a large external dataset. RRAE, LM-A, LM-S, MN, and TC are weakly supervised, utilizing video-level annotations and training on large external web-crawled video datasets. MT is unsupervised but uses a large external dataset for pretraining. In comparison, without any weak supervision or external dataset, our approach not only outperforms the prior methods but also improves the state-of-the-art by almost 3\%. This indicates the superiority of our unsupervised highlight detection model trained using pseudo-highlights based on audio and visual recurrence.

\setlength{\tabcolsep}{2.2pt}
\begin{table}[htb]
\centering
\small
\begin{tabular}{@{}lcccccccc@{}}
\toprule
Method    & RRAE & LM-A & LM-S & MN & TC & MT & CHD & Ours \\ \midrule
Ext. data & \cmark           & \cmark             & \cmark                        & \cmark                      & \cmark                      & \cmark                      & \xmark                       & \xmark                       \\
mAP      & 38.30             & 50.50        & 56.40                     & 61.38                   & 62.97                   & 65.10                   & 65.39                    & \textbf{68.30}            \\ \bottomrule
\end{tabular}
\caption{Highlight detection results on YouTube Highlights. Our approach outperforms all prior unsupervised and weakly supervised methods, including those that rely on external data.}

\label{table:comparison_YouTube}
\end{table}

{\textbf{TVSum:}} In Table \ref{table:comparison_tvsum_1}, we compare the performance of our approach on TVSum with prior unsupervised and weakly supervised methods that do not rely on an external dataset for training. Once again, we achieve state-of-the-art performance on this dataset, outperforming the best prior method CHD by a large margin of about 8\%. Moreover, in Table \ref{table:comparison_tvsum_2}, we compare our method with prior weakly supervised methods that rely on a large external dataset of web-crawled videos for training. Even without the advantage of using additional external data, our method outperforms three of these methods. Notably, we outperform LM-A and LM-S which are trained with 10 million Instagram videos. 

\setlength{\tabcolsep}{4pt}
\begin{table}[h]
\centering
\small
\begin{tabular}{@{}lcccccc@{}}
\toprule
Method    & MBF  & CVS  & SG   & DSN  & CHD  & Ours \\ \midrule
Ext. data & \xmark    & \xmark    & \xmark    & \xmark    & \xmark    & \xmark    \\
top-5 mAP & 34.50 & 37.20 & 46.20 & 42.40 & 52.76 & \textbf{60.34} \\ \bottomrule
\end{tabular}
\caption{Highlight detection results on TVSum. We compare with existing unsupervised and weakly supervised methods that do not rely on external data. Our method surpasses the prior state-of-the-art method by a large margin of about 8\%.}
\label{table:comparison_tvsum_1}
\end{table}

\setlength{\tabcolsep}{3pt}
\begin{table}[h]
\centering
\small
\begin{tabular}{@{}lccccccc@{}}
\toprule
Method    & VESD & LM-A & LM-S & MN   & TC   & MT   & Ours \\ \midrule
Ext. data & \cmark    & \cmark & \cmark    & \cmark    & \cmark    & \cmark   & \xmark    \\
top-5 mAP & 42.30 & 52.40 & 56.30 & 73.24 & 76.82 & 78.30 & 60.34 \\ \bottomrule
\end{tabular}
\caption{We compare with prior methods on TVSum that utilize external data. Even without the advantage of external data, our method outperforms three of these methods.}
\label{table:comparison_tvsum_2}
\end{table}

{\textbf{QVHighlights:}} In Table \ref{table:comparison_qvhighlights}, we evaluate our method on the QVHighlights dataset. Note that these results are on its \textit{test} split, which requires submission of predictions on their evaluation server. Our approach significantly outperforms both of the prior unsupervised methods BT and CHD which, similar to our work, do not utilize query information by a significant margin. BT selects key frames based on their aesthetic quality and relevance to the content of the video, while CHD is a recent unsupervised method that utilizes a contrastive objective to train its model. CHD does not evaluate on QVHighlights, so we implement and evaluate their method on this dataset to compare with our approach. 
\setlength{\tabcolsep}{8pt}
\begin{table}[htb]
\centering
\small
\begin{tabular}{@{}lccc@{}}
\toprule 
Method    & BT & CHD  &  Ours     \\ \midrule
mAP & 14.36    &  15.82     & \textbf{18.38}         \\
HIT@1 & 20.88 & 17.10 & \textbf{24.71} \\ \bottomrule
\end{tabular}
\caption{Highlight detection results on the QVHighlights \textit{test} split from the evaluation server. We compare with unsupervised methods that do not consider user query annotations for the videos. Our method achieves state-of-the-art performance on both metrics.}
\label{table:comparison_qvhighlights}
\end{table}

\begin{figure*}[htb] 
\centering
\includegraphics[width=0.75\textwidth]{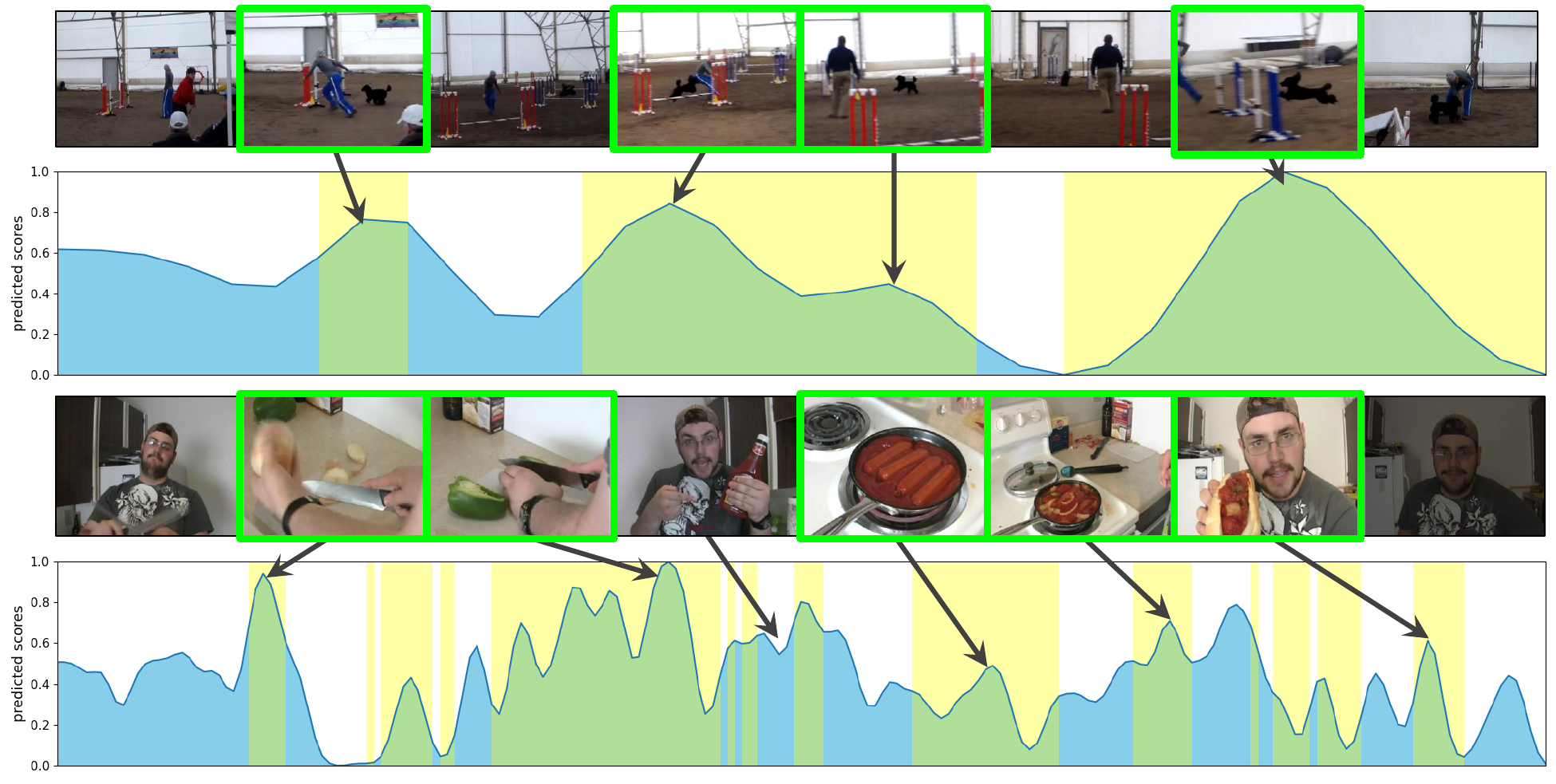} 
\caption{Qualitative results. We show the highlight clips (in green) along with the predicted scores of our method (in blue), with the ground truth highlights regions (indicated in yellow). For the \textit{dog show} video from the YouTube dataset (top), our method correctly picks clips with interesting acrobatic movements such as jumping over obstacles. For the video depicting \textit{making sandwich} from TVSum (bottom), our method accurately detects the highlight moments, which mostly consist of close-up shots of food and important stages of cooking.}
\label{fig:qual}
\end{figure*}

\textbf{Qualitative Results:} We present some qualitative results of our method in Figure \ref{fig:qual}. The video on the top row from the YouTube dataset depicts a \textit{dog show} and the highlight mostly consists of acrobatics such as jumping over obstacles. The bottom one from TVSum depicts \textit{making sandwich}, with mostly close-up shots of food indicated as highlights. In both, our model correctly identifies the highlights.


\setlength{\tabcolsep}{2pt}
\subsection{Ablation Studies}
We conduct extensive ablation studies to analyze the relative impact of each component of our approach. For QVHighlights, we follow prior work \cite{lei_moment_detr, liu2022umt} and utilize its \textit{val} split due to limited number of submissions to the evaluation server. For TVSum and YouTube, we use the \textit{test} split. We define the following baselines for our experiments.

\begin{itemize}[ nosep, wide=0pt, itemindent = 10pt]
  \item \textbf{A, V, and AV:} A and V denote unimodal  models with a single self-attention layer that is trained on only audio and visual features, respectively. AV denotes the audio-visual highlight detection model in our proposed method.
  \item \textbf{A-PH, V-PH, and AV-PH:} A-PH refers to audio-visual highlight detection model that is trained using only audio pseudo-highlights. V-PH indicates the model trained using only visual pseudo-highlights. Finally, AV-PH, trained using audio-visual pseudo-highlights, is the proposed model.
\end{itemize}

{\textbf{Pseudo-highlight Variations:}} We examine the relative impact of each modality and their combination for pseudo-highlight generation in Table \ref{table:ablation_pseudohighlights}. On all three datasets, the audio-visual highlight detection model trained using audio pseudo-highlights (A-PH) alone performs better than visual pseudo-highlights (V-PH). This indicates that recurring moments in the audio can provide strong cues for detecting highlights, emphasizing the significance of audio modality. The results also show combined  audio-visual pseudo-highlights (AV-PH) are more effective than using either modality alone, as these two modalities can contain complementary information about potential highlight moments.

\begin{table}[h]
\centering
\small
\begin{tabular}{@{}lcccc@{}}
\toprule
\multicolumn{1}{c}{\multirow{2}{*}{Method}}
& TVSum           & YouTube         & \multicolumn{2}{c}{QVHighlights \it{val}} \\ \cmidrule(lr){2-2} \cmidrule(lr){3-3} \cmidrule(lr){4-5}
\multicolumn{1}{c}{}                        & top-5 mAP       & mAP             & mAP        & HIT@1     \\ \midrule
AV(V-PH)   & 53.35          & 62.76          & 17.58           & 20.19          \\
AV(A-PH)   & 58.34          & 66.36          & 17.81           & 23.10           \\
AV(AV-PH) (Ours)   & \textbf{60.34} & \textbf{68.30} & \textbf{18.41}  & \textbf{26.26} \\ \bottomrule
\end{tabular}
\caption{Ablation study for investigating the relative impact of generating pseudo-highlights from audio and visual modality.}
\label{table:ablation_pseudohighlights}
\end{table}

{\textbf{Unimodal vs. Bimodal Setting:}} Next, we analyze the effectiveness of our approach in unimodal settings, where only one modality is available during both pseudo-highlight generation and model training. We train the visual unimodal model (V) using visual pseudo-highlights and the audio unimodal model (A) using audio pseudo-highlights. We compare their performance with our approach in Table \ref{table:ablation_unimodal}. While both unimodal settings perform competitively with prior methods, learning from both visual and audio features (i.e., our method) together significantly boosts performance.

\begin{table}[htb]
\centering
\small
\begin{tabular}{@{}lcccc@{}}
\toprule
\multicolumn{1}{c}{\multirow{2}{*}{Method}} & TVSum           & YouTube         & \multicolumn{2}{c}{QVHighlights \it{val}} \\ \cmidrule(lr){2-2} \cmidrule(lr){3-3} \cmidrule(lr){4-5}
\multicolumn{1}{c}{}                        & top-5 mAP       & mAP             & mAP        & HIT@1     \\ \midrule
V(V-PH)   & 55.02          & 59.98          & 16.38           & 21.35          \\
A(A-PH)   & 55.45          & 62.07          & 17.37           & 22.06           \\
AV(AV-PH) (Ours)                              & \textbf{60.34} & \textbf{68.30}  & \textbf{18.41}  & \textbf{26.26} \\ \bottomrule
\end{tabular}
\caption{Ablation on unimodal vs. bimodal setting. We analyze the relative contribution of each modality by using only one modality during both pseudo-highlight generation and model training.}
\label{table:ablation_unimodal}
\end{table}

{\textbf{Similarity Metrics:}} To analyze the effectiveness of cosine similarity in our audio and visual pseudo-highlight generation method, we replace it with another popular similarity measure, Pearson's Correlation Coefficient (PCC) \cite{cohen2009pearson}. While cosine similarity only considers the similarity of orientation of two feature vectors, PCC considers the linear relationship between features, incorporating both orientation and magnitude. Table \ref{table:similarity} indicates that cosine similarity is more effective in capturing the similarity between audio or visual features for pseudo-highlight generation.

\begin{table}[htb]
\centering
\small
\begin{tabular}{@{}lcccc@{}}
\toprule
\multicolumn{1}{c}{\multirow{2}{*}{Method}} & TVSum           & YouTube         & \multicolumn{2}{c}{QVHighlights \it{val}} \\ \cmidrule(lr){2-2} \cmidrule(lr){3-3} \cmidrule(lr){4-5}
\multicolumn{1}{c}{}                        & top-5 mAP       & mAP             & mAP        & HIT@1     \\ \midrule
PCC  & 52.59    & 58.71    & 17.60    & 23.61  \\
Cosine (Ours)  & \textbf{60.34} & \textbf{68.30}  & \textbf{18.41}  & \textbf{26.26} \\ \bottomrule
\end{tabular}
\caption{Ablation on similarity metrics in pseudo-highlight generation. We replace the cosine similarity in our method with Pearson's correlation coefficient (PCC).}
\label{table:similarity}
\end{table}

{\textbf{Audio-Visual Fusion Techniques:}} In Table \ref{table:ablation_fusion}, we explore various methods of combining audio and visual features during training, as done in previous work \cite{badamdorj2021joint}. We compare our method with SA\textsuperscript{early} and SA\textsuperscript{late}. In SA\textsuperscript{early}, both audio and visual features are first concatenated and then fed into a single self-attention layer. However, in SA\textsuperscript{late}, each modality is initially processed using separate self-attention layers in a two-stream fashion, and the output features are concatenated for highlight detection. Our fusion scheme, consisting of a bimodal attention module, significantly outperforms these alternative fusion schemes due to its ability to better capture complex cross-modal interactions.

\setlength{\tabcolsep}{2pt}
\begin{table}[htb]
\centering
\small
\begin{tabular}{@{}lcccc@{}}
\toprule
\multicolumn{1}{c}{\multirow{2}{*}{Method}} & TVSum           & YouTube         & \multicolumn{2}{c}{QVHighlights \it{val}} \\ \cmidrule(lr){2-2} \cmidrule(lr){3-3} \cmidrule(lr){4-5}
\multicolumn{1}{c}{}                        & top-5 mAP       & mAP             & mAP        & HIT@1     \\ \midrule
SA\textsuperscript{early}   & 56.68          & 61.23          & 16.25           & 17.87          \\
SA\textsuperscript{late}   & 53.74          & 65.61          & 17.78           & 23.87           \\
AV(AV-PH) (Ours)  & \textbf{60.34} & \textbf{68.30} & \textbf{18.41}  & \textbf{26.26} \\ \bottomrule
\end{tabular}
\caption{Effect of different audio-visual fusion strategies for combining audio and visual features during training.}
\label{table:ablation_fusion}
\end{table}

\textbf{Impact of different number of clusters, $K$:} In Table \ref{table:ablation_cluster_number}, we compare the performance of our method for different values of number of clusters, $K$. Results show a performance drop for non-optimal values of $K$, while the optimal values chosen by our method in Sec. \ref{subsec:gpc} yield best results.

\setlength{\tabcolsep}{6pt}
\begin{table}[htb]
\centering
\small
\begin{tabular}{@{}lccc@{}}
\toprule
Number of clusters    & 6 & 7 & 8 \\ \midrule
TVSum                  & 52.84       &  54.76    &	\textbf{60.34}\\ 
YouTube             &   \textbf{68.30}	& 63.96	  &   63.56\\ 
QVHighlights       &  17.80	 &  \textbf{18.41}	&   18.02 \\ 
\bottomrule
\end{tabular}
\caption{Ablation on the different number of clusters, $K$. The optimal values of $K$ selected by our method yield best results.}
\label{table:ablation_cluster_number}
\end{table}

{\textbf{Comparison with Supervised Setting:}} In Table \ref{table:ablation_groundtruth}, we compute the highlight detection performance when real ground-truths are used for training our model instead of the audio-visual pseudo-highlights. This denotes the fully supervised version of our model. Interestingly, our unsupervised method demonstrates strong performance compared to its supervised counterpart. On YouTube, our method underperforms by only 2\% compared to the supervised model. This showcases the effectiveness of the proposed pseudo-highlight mechanism as an alternative supervisory signal for highlight detection if manual annotations are not available.

\begin{table}[htb]
\small
\begin{tabular}{@{}lcccc@{}}
\toprule
\multicolumn{1}{c}{\multirow{2}{*}{Method}} & TVSum           & YouTube         & \multicolumn{2}{c}{QVHighlights \it{val}} \\ \cmidrule(lr){2-2} \cmidrule(lr){3-3} \cmidrule(lr){4-5}
\multicolumn{1}{c}{}  & top-5 mAP       & mAP             & mAP        & HIT@1     \\ \midrule
AV(Supervised) &68.41 &70.18    &23.99    &32.32  \\ 
AV(AV-PH) (Ours) &60.34 &68.30 &18.41  &26.26 \\ \bottomrule
\end{tabular}
\caption{Comparison with the supervised version of our model.}
\label{table:ablation_groundtruth}
\end{table}

{\textbf{Real Categories vs. Pseudo-categories:}} To analyze the quality of our generated pseudo-categories for audio-visual pseudo-highlight generation, we replace them with real category information in the datasets. Note that the QVHighlights dataset does not have category information, so we limit our comparison and analysis to the YouTube and TVSum datasets. Table \ref{table:ablation_category} shows that we obtain better performance using our generated pseudo-categories in comparison to the real categories. We argue that manually annotated category labels can contain ambiguities. Through our clustering-based pseudo-categories, we are able to group semantically related videos that may have been manually labeled as different categories. As a result, our clustering-based pseudo-categories enable us to generate better audio-visual pseudo-highlights compared to using real categories.

\begin{table}[htb]
\centering
\small
\begin{tabular}{@{}lcccc@{}}
\toprule
\multicolumn{1}{c}{Method} & TVSum           & YouTube   \\ \midrule
Real categories  & 56.37    & 64.28 \\
Pseudo-categories (Ours)                              & \textbf{60.34} & \textbf{68.30}  \\ \bottomrule
\end{tabular}
\caption{Comparison of using pseudo-categories and real categories in our method. Generating audio-visual pseudo-highlights using our clustering-based pseudo-categories performs better.}
\label{table:ablation_category}
\end{table}

{\textbf{How does the amount of pseudo-highlights impact the performance?}} We reduce the amount of training data in each dataset to 25\%, 50\%, and 75\%, yielding different amounts of audio-visual pseudo-highlights. Table \ref{table:ablation_amount} shows that training with more pseudo-highlights improves performance. This suggests that using more pseudo-highlights for supervision is beneficial for video highlight detection.

\setlength{\tabcolsep}{2pt}
\begin{table}[htb]
\centering
\small
\begin{tabular}{@{}lcccc@{}}
\toprule
\multicolumn{1}{c}{\multirow{2}{*}{}} & TVSum           & YouTube         & \multicolumn{2}{c}{QVHighlights \it{val}} \\ \cmidrule(lr){2-2} \cmidrule(lr){3-3} \cmidrule(lr){4-5}
\multicolumn{1}{c}{}                        & top-5 mAP       & mAP             & mAP        & HIT@1     \\ \midrule
25\%   & 47.31          & 60.55          & 17.63           & 23.23         \\
50\%   & 55.37          & 62.16          & 18.13           & 23.68          \\
75\%   & 54.64          & 64.25          & 18.16           & 24.26          \\
100\% (Ours)  & \textbf{60.34} & \textbf{68.30} & \textbf{18.41}  & \textbf{26.26} \\ \bottomrule
\end{tabular}
\caption{Impact of the amount of pseudo-highlights on the performance of our model. More pseudo-highlights yield better results.}
\label{table:ablation_amount}
\end{table}

{\textbf{Audio-visual pseudo-highlight as supervision:}} We conduct an experiment where we directly use the audio-visual pseudo-highlight scores of each video as predictions and compare them with predictions of our highlight detection model trained using audio-visual pseudo-highlights. Concretely, we first assign each test video to its corresponding pseudo-category by finding the nearest cluster centroid in the clustering model learned on train data in Sec. \ref{subsec:gpc}. Then, we obtain the audio-visual pseudo-highlight scores of each test video by comparing them with all training videos of the same pseudo-category. We directly use audio-visual pseudo-highlight scores of a test video as its highlight predictions. In Table \ref{table:evaluation_pseudohighlights}, we compare these highlight predictions with highlights predicted by our trained model. Despite potential noise in pseudo-highlight supervision, our model is able to learn from it during training, indicating its worth and benefits, and achieves superior performance compared to using pseudo-highlight directly as predictions.

\begin{table}[htb]
\centering
\small
\begin{tabular}{@{}lcccc@{}}
\toprule
\multicolumn{1}{c}{\multirow{2}{*}{Method}} & TVSum           & YouTube         & \multicolumn{2}{c}{QVHighlights \it{val}} \\ \cmidrule(lr){2-2} \cmidrule(lr){3-3} \cmidrule(lr){4-5}
\multicolumn{1}{c}{}                        & top-5 mAP       & mAP             & mAP        & HIT@1     \\ \midrule
pseudo-highlights & 52.59    & 58.71    & 17.60    & 23.61  \\ 
AV(AV-PH) (Ours) & \textbf{60.34} & \textbf{68.30}  & \textbf{18.41}  & \textbf{26.26} \\ \bottomrule
\end{tabular}
\caption{Evaluating the quality of audio-visual pseudo-highlights as supervision for highlight detection. We compare the predictions of our model trained using pseudo-highlights (second row) with the pseudo-highlight scores used directly as predictions (first row).}
\label{table:evaluation_pseudohighlights}
\end{table}

%% file: texs/conclusion.tex
\section{Conclusion}\label{sec:conclude}
We introduce a novel unsupervised audio-visual highlight detection framework. Our core idea is based on the premise that videos of similar categories tend to contain key moments that are repetitive in both audio and visual modalities across multiple videos. Leveraging this observation, we construct audio-visual pseudo-highlights to train our model. Extensive experiments showcase the effectiveness of our framework and reveal that cues from the often overlooked yet informative audio modality, when coupled with the visual modality, lead to a significant improvement in unsupervised learning of video highlight detection.

\textbf{Acknowledgements}: Zahidul Islam and Mrigank Rochan acknowledge the support of the University of Saskatchewan and the Natural Sciences and Engineering Research Council of Canada (NSERC).